\pdfoutput=1

\documentclass[11pt]{article}

\usepackage[]{acl}

\usepackage{listings}
\usepackage{times}
\usepackage{latexsym}
\usepackage{algorithm}
\usepackage{algorithmicx}
\usepackage{algpseudocode}
\usepackage{float}

\usepackage[T1]{fontenc}

\usepackage[utf8]{inputenc}

\usepackage{microtype}

\usepackage{inconsolata}
\usepackage{tabularray}
\usepackage{todonotes}
\UseTblrLibrary{booktabs}
\usepackage{subcaption}
\graphicspath{{./figures}}

\title{Human Still Wins over LLM: An Empirical Study of Active Learning on Domain-Specific Annotation Tasks}

\author{Yuxuan Lu \\
  Northeastern University \\\And
  Bingsheng Yao \\ Rensselaer Polytechnic Institute \\\And
  Shao Zhang \\ Shanghai Jiao Tong \\ University \\\AND
  Yun Wang \\ Microsoft Research Asia\\\And
  Peng Zhang \\ Fudan University \\\And
  Tun Lu \\ Fudan University \\\AND
  Toby Jia-Jun Li \\ University of Notre Dame \\\And
  Dakuo Wang \thanks{~~Corresponding Author: \texttt{d.wang@northeastern.edu} .}  \\ Northeastern University }

\begin{document}
\maketitle
\begin{abstract}

Large Language Models (LLMs) have demonstrated considerable advances, and several claims have been made about their exceeding human performance. However, in real-world tasks, domain knowledge is often required. Low-resource learning methods like Active Learning (AL) have been proposed to tackle the cost of domain expert annotation, raising this question: \textbf{Can LLMs surpass compact models trained with expert annotations in domain-specific tasks?}
In this work, we conduct an empirical experiment on \textbf{four} datasets from \textbf{three} different domains comparing SOTA LLMs with small models trained on expert annotations with AL. We found that small models can outperform GPT-3.5 with a few hundreds of labeled data, and they achieve higher or similar performance with GPT-4 despite that they are hundreds time smaller. Based on these findings, we posit that LLM predictions can be used as a warmup method in real-world applications and human experts remain indispensable in tasks involving data annotation driven by domain-specific knowledge.

\end{abstract}

\section{Introduction}





Instructional-finetuned Large Language Models~\cite{chungScalingInstructionFinetunedLanguage2022, weiFinetunedLanguageModels2022,ouyangTrainingLanguageModels2022} (LLMs) have shown significant advances in their zero-shot and few-shot capabilities, i.e., in-context learning~\cite{brownLanguageModelsAre2020}, on reasoning, arithmetic, and comprehension. 
Several recent works claim that LLMs can even outperform human crowdworkers for several data annotation tasks, such as Text Classification \cite{gilardiChatGPTOutperformsCrowdWorkers2023}, Evaluation \cite{chiangCanLargeLanguage2023, liuGEvalNLGEvaluation2023}, etc. 

In reality, however, a multitude of real-world tasks necessitates the domain knowledge of human experts in specific domains, such as doctors making diagnoses or lawyers reviewing contracts. 
Traditionally, such human experts have been the singular source of high-quality data in their professional fields, crucial for achieving downstream goals, such as fine-tuning a language model with those high-quality annotations. 
However, experts' annotations are frequently associated with substantial costs, limited access, and numerous challenges~\cite{wu2022survey}.

In response to these challenges, the research community has put forward an array of methods to reduce the annotation cost, i.e., learning with low resources. 
Active learning (AL) \cite{prince2004does, settles2009active, zhang2022survey} is one such popular framework to iteratively sample a few representative data, query human experts' annotations, and then train the models. 
AL has been demonstrated effective for various low-resource scenarios~\cite{sharmaActiveLearningRationales2015, yaoLabelsEmpoweringHuman2023}.
Thus, we ask the following question: \textbf{Do LLMs necessarily beat small models trained with limited experts' annotations in domain-specific tasks?}

We hypothesize that small language models can quickly learn domain knowledge from a limited set of labeled data with low-resource learning techniques, such as Active Learning, whereas LLMs may not perform well due to their lack of domain-specific knowledge or fine-tuning.
In this work, we conduct an empirical study with Active Learning simulations on four datasets (BioMRC, ContractNLI, Unfair\_TOS, FairytaleQA) of three types of tasks (Multiple Choice, Classification, and Question Generation) from three different domains (Biomedicine, Law, and Education).
We probe the best-performing state-of-the-art (SOTA) LLMs and compare them with a much smaller model fine-tuned with different AL techniques.
Our results not only demonstrate the effectiveness of leveraging AL for finetuning small models with a very small amount of data in different professional domains but also illustrate the irreplaceability of human experts' annotations, despite very little demand by efficient Active Learning strategies. 



\section{Related Works}

\subsection{Large Language Models for Data Annotation}
Since the emergence of Large Language Models (LLMs) \cite{brownLanguageModelsAre2020,OpenAI2023GPT4TR,touvronLLaMAOpenEfficient2023,touvronLlamaOpenFoundation2023}, they have been extensively employed in various data annotation tasks, such as Text Classification \cite{gilardiChatGPTOutperformsCrowdWorkers2023, kuzmanChatGPTBeginningEnd2023, tornbergChatGPT4OutperformsExperts2023}, Text Quality Evaluation \cite{chiangCanLargeLanguage2023, liuGEvalNLGEvaluation2023}, and there are several reports suggesting that LLMs outperform crowd workers \cite{gilardiChatGPTOutperformsCrowdWorkers2023} and even domain experts \cite{tornbergChatGPT4OutperformsExperts2023} in certain tasks. However, these tasks usually have a relatively low requirement on domain knowledge, and the ability of LLMs to handle data annotation in tasks that require high levels of specialized knowledge remains an area yet to be investigated.

\subsection{Active Learning for Data Annotation}
The expensive nature of expert-generated annotations has sparked interest in methods that can effectively learn from a constrained number of labeled samples. Active Learning (AL)~\citep{sharmaActiveLearningRationales2015, shenDeepActiveLearning2017, ash2019deep, teso2019explanatory, kasaiLowresourceDeepEntity2019, zhang-etal-2022-allsh, yaoLabelsEmpoweringHuman2023} is a cyclical process that involves: 
1) selecting examples from an unlabeled data repository (utilizing AL selection strategies) to be labeled by human annotators, 
2) training the model with the newly labeled data, and 
3) assessing the tuned model's performance.

A few AL surveys~\cite{settles2009active, olsson2009literature, fu2013survey, schroder2020survey, ren2021survey} of sampling strategies provide two high-level selection concepts: data diversity-based strategies and model uncertainty-based strategies. In our work, both types of methods are experimented with.

\section{Empirical Study}
Our empirical study probes SOTA LLMs' best-performing setup off-the-shelf, including GPT-3.5 and GPT-4~\cite{OpenAI2023GPT4TR}, and compares them with a traditional language model, which is T5-base~\cite{raffel2020exploring}, with various AL techniques. 
It is worth mentioning that T5-base comprised only $220$ million parameters, whereas GPT-4 is rumored to surpass trillion-level parameters.

\subsection{Datasets}

\begin{table}
  \center
  \small
  \begin{booktabs}{
    width=\linewidth,
    colspec={ccX[1,c]c},
    hspan=minimal,
    cells={m}
  }
    \toprule
     Dataset & Domain & Task & {\#Test \\ Samples} \\
    \midrule
    BioMRC & Biomed. & {Multiple \\ Choice} & 6250 \\
    Unfair\_tos & Law & Classification & 1620 \\
    ContractNLI & Law & NLI & 1991 \\
    FairytaleQA & Edu. & {Question \\ Generation} & 1007 \\
    \bottomrule
  \end{booktabs}
  \caption{Datasets involved in our empirical study.}
  \label{tab:dataset}
  \vspace{-\baselineskip}
\end{table}
As our primary target is to evaluate models' performance in tasks that necessitate a profound understanding of domain-specific knowledge, we chose the BioMRC \cite{pappasBioMRCDatasetBiomedical2020}, Unfair\_tos \cite{lippiCLAUDETTEAutomatedDetector2019}, ContractNLI \cite{koreedaContractNLIDatasetDocumentlevel2021} and FairytaleQA \cite{xuFantasticQuestionsWhere2022} respectively on Multiple Choice, Classification, NLI, and Question Generation tasks to evaluate our system's ability. The details for these datasets can be found in table \ref{tab:dataset}.

\subsection{Evaluation Setup}
We use T5 \cite{raffel2020exploring} 
as the representative of pre-trained small models and use GPT-3.5\footnote{GPT-3.5-0613} and GPT-4 \cite{OpenAI2023GPT4TR}\footnote{GPT-4-0613} as representatives for SOTA LLMs. 
For each task, we utilize the same instruction prompts which can be found in Appendix \ref{sec:prompt}.

\subsubsection{Large Language Models}
In order to establish the best-performing benchmarks for LLMs, we experiment with zero-shot and few-shot prompting (1, 3, and 10 shots) on GPT-3.5, and run GPT-4 under the same setting. 
We put the task description prompt as the System message, the question from the few shot examples as the User message, and the answers from the few shot examples as the Assistant message.

\begin{figure*}[t]
    \centering
     \begin{subfigure}[b]{0.43\linewidth}
         \centering
         \includegraphics[width=\linewidth]{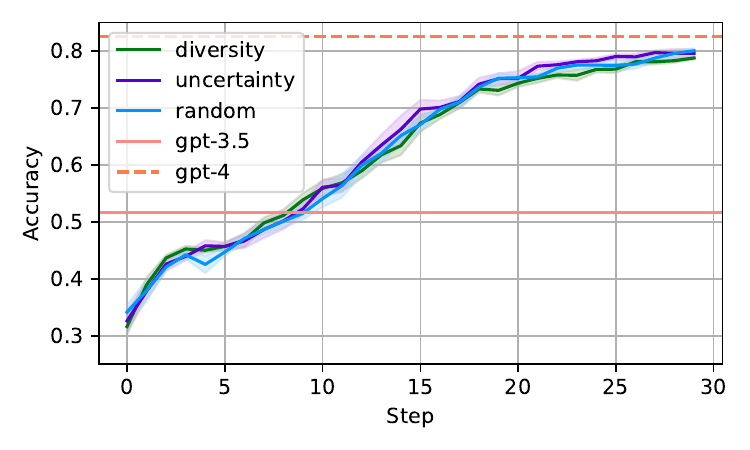}
         \caption{BioMRC}
         \label{fig:results-biomrc}
     \end{subfigure}
     \begin{subfigure}[b]{0.43\linewidth}
         \centering
         \includegraphics[width=\linewidth]{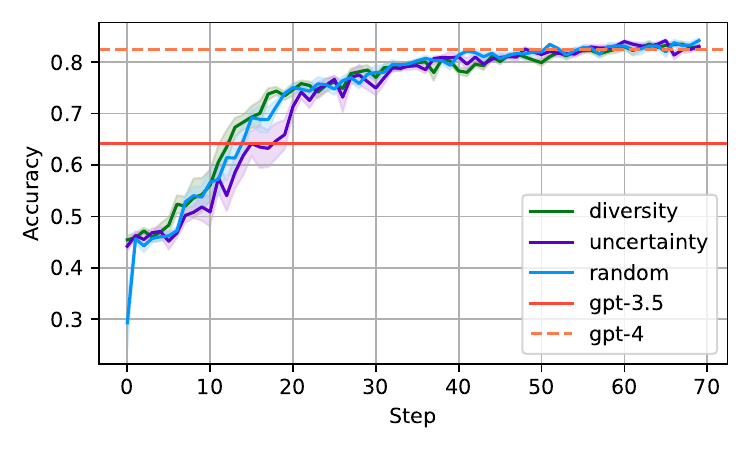}
         \caption{ContractNLI}
         \label{fig:results-contractnli}
     \end{subfigure}
     \begin{subfigure}[b]{0.43\linewidth}
         \centering
         \includegraphics[width=\linewidth]{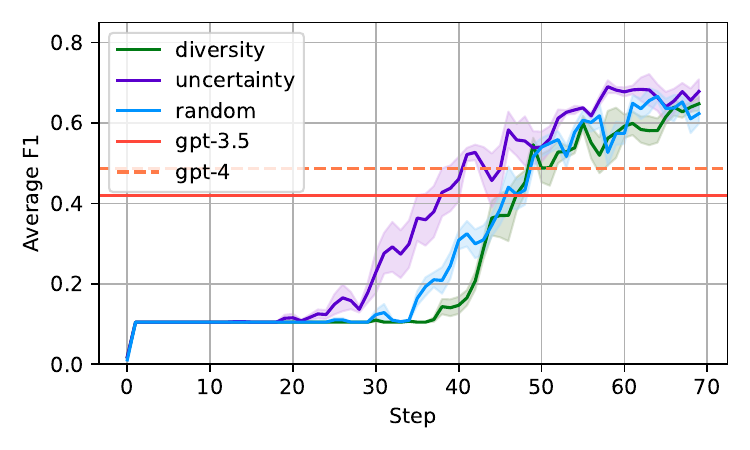}
         \caption{Unfair\_TOS}
         \label{fig:results-unfairtos}
     \end{subfigure}
     \begin{subfigure}[b]{0.43\linewidth}
         \centering
         \includegraphics[width=\linewidth]{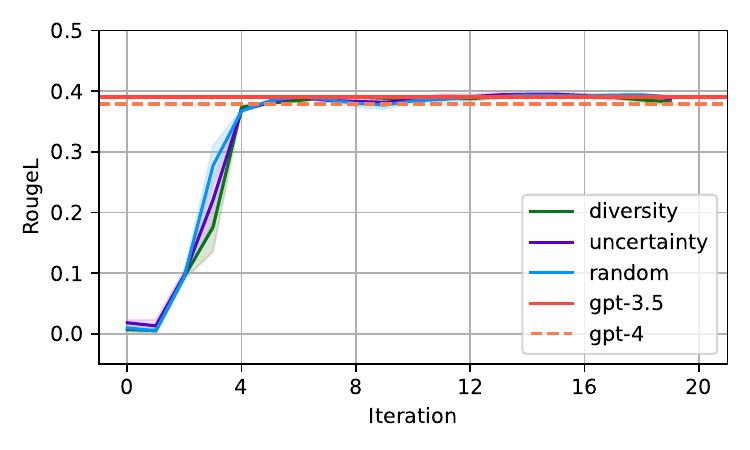}
         \caption{FairytaleQA}
         \label{fig:results-fairytaleqa}
     \end{subfigure}
    \caption{Emipirical study results. The horizontal line represents the best performance yielded by LLMs. We report the mean value (line) and standard error (colored shadow area) in 10 runs for each setting.}
    \label{fig:results}
\end{figure*}

\begin{algorithm}[t]
    \caption{Data Diversity-based data sampling}
    \label{alg:data}
    \begin{algorithmic}[1]
        \Function{select}{$D_{train}, D_{prev}, N$}
            \State $D_{train}$: unlabeled data in the training split
            \State $D_{prev}$: previously selected data
            \State $N$: number of data needed
            \State $Scores \gets 0$
            \For{$d_i \in D_{train}$}
                \State $Scores_{i} \gets \frac{1}{|D_{prev}|} \times \sum_{d_p \in D_{prev}} Similarity(d_i, d_p)$
            \EndFor
            \State $id \gets \mathrm{argsort}(Scores)$
            \State $step \gets \frac{|D_{train}|}{N}$
            \State $D_{iter} \gets \{ D_{train, id_i} | i \parallel step \}$
            \State \textbf{return} $D_{iter}, D_{train} - D_{iter}$
        \EndFunction
    \end{algorithmic}
\end{algorithm}

\begin{algorithm}[t]
    \caption{Uncertainty based data sampling}
    \label{alg:u}
    \begin{algorithmic}[1]
        \Function{select}{$D_{train}, N$}
            \State $D_{train}$: unlabeled data in the training split
            \State $N$: number of data needed
            \State $Scores \gets 0$
            \For{$d_i \in D_{train}$}
                \State $Scores_{i} \gets \mathrm{Uncertainty}(d_i)$
            \EndFor
            \State $id \gets \mathrm{argsort}(Scores)$
            \State \textbf{return} $id_{ \le N}, id_{> N}$
        \EndFunction
    \end{algorithmic}
\end{algorithm}

\subsubsection{Active Learning}
Active Learning aims to select the most needed data to annotate in each iteration to improve label efficiency.  Following common practice, we run active learning simulations, where in each iteration, we sample 16 data samples with the specified strategy and train the model on that batch of data, then evaluate our model on test split afterward. We run the active learning loop 10 times with the same setting and report the mean and standard error. Model evaluations are performed on the test splits of the datasets following each training iteration. For BioMRC and ContractNLI, testing is carried out on a sampled subset of 1,000 test instances. Both data sampling process is shown in Algo. \ref{alg:data} and \ref{alg:u}.

\begin{itemize}
    \item \textbf{Data diversity-based Active Learning}. The objective of data diversity-based Active Learning \cite{schroderRevisitingUncertaintybasedQuery2022} is to identify the most representative and diverse datasets for subsequent training batches. Initially, we randomize our selection of data from the pool of available training data. In each subsequent iteration, we embed each unlabeled instance in the pool using SentenceTransformer \cite{wangMiniLMDeepSelfAttention2020} and then compute its average cosine similarity with all samples selected previously. We then uniformly select (with the same step) training data from the ranked list of average similarity scores. This balanced approach ensures the model benefits from exposure to familiar and novel samples.

    \item \textbf{Uncertainty-based Active Learning} \cite{senerActiveLearningConvolutional2018} aspires to identify the most uncertain samples for subsequent training iterations. Within each iteration, the model operates on a randomly sampled subset of the training data, computing the model's logits and locating the samples holding the minimal average probability on the highest-ranked tokens. Thus, the most confident samples for the model to train can be discerned. 
\end{itemize}

\subsection{Evaluation Metrics}

For BioMRC and ContractNLI, we use average Accuracy as the evaluation metric. For Unfair\_TOS, as it's an unbalanced multiple-class classification task, we compute F1 score separately on each label and report the average F1 score to avoid the influence of unbalanced labels.

\section{Evaluation Results}

\subsection{Comparision between Large Language Models and Active Learning Models}

The results are shown in Fig. \ref{fig:results}. The horizontal line symbolizes the best performance for both GPT-3.5 and GPT-4 respectively. \textbf{On all 4 datasets, all models quickly outperform GPT-3.5 after learning from a few data samples.} After accumulating a total of several hundreds of data points on most of the tasks, the models reach a performance that is compatible with GPT-4.

On BioMRC and ContractNLI, GPT-4 presents exceptional performance, although it may have trained on these datasets. Regardless, our fine-tuned models manage to achieve performance levels that are compatible with GPT-4, despite having hundreds of times fewer parameters and requiring significantly less computational power.

In Unfair\_TOS, approximately 90\% of test samples are labeled ``None'', and prior to the 20th iteration, all models merely output ``None'' regardless of the input. Following this, models using the uncertainty-based approach quickly learned to recognize other labels, followed by those employing other strategies. These models rapidly outperformed both GPT-3.5 and GPT-4.

On FairytaleQA, considering its generative essence, RougeL may not be an accurate representation of their performance \cite{xuFantasticQuestionsWhere2022}. Still, our models achieved similar performance with GPT-3.5 and GPT-4.

\subsection{Comparison between different Active Learning strategies}

Different active learning strategies yield varying outcomes across different tasks. The uncertainty-based data sampling approach proved more effective on BioMRC and Unfair\_TOS, while the diversity-based and random strategies demonstrated superior performance on ContractNLI and FairytaleQA, respectively.




\section{Observations and Discussion}

GPT-4 yields impressive performance on all these tasks. Interestingly, the performance gap between GPT-3.5 and GPT-4 on BioMRC and ContractNLI is significantly higher than that on other datasets. We suspect that GPT-4 may have seen these two datasets during their training process.

The performance disparity between diversity-based and uncertainty-based methods appears to correlate with label distributions. For instance, in ContractNLI, the ratio of three labels is approximately 5:4:1, whereas in Unfair\_TOS, 90\% of test samples are labeled as ``None''. We speculate that opting for data samples evenly across different similarity groups could contribute to a higher selection of data labeled as ``None''.

\section{Conclusion}

While LLMs such as GPT-4 have been endorsed as simulated data annotators, proficiently outpacing human annotators in data labeling tasks, our paper presents a benchmarking experiment evaluating the Active Learning-aided method. This method involves human domain experts annotating 4 domain-knowledge-dependent datasets on 3 domains and comparing them against GPT-4 and GPT-3.5 few-shot baseline. 

The results from our experiment demonstrate that AL-assisted expert annotation can rapidly achieve or exceed the performance of a GPT-simulated annotator after a few initial iterations, and eventually saturates at a comparable performance level. Based on these findings, we posit that LLM predictions can be used as a warmup method in real-world applications and human experts remain indispensable in tasks involving data annotation driven by domain-specific knowledge.

\section{Limitations}

There are certain limitations in our work: 1) Our experiments with active learning solely utilize T5 models, where the performance of other models remains to be explored. 2) We primarily engage in model comparisons through automated metrics. However, these may not necessarily provide an accurate representation of a model's performance, particularly for Question Generation tasks. Therefore, human evaluation of these datasets should be conducted for a more comprehensive assessment.
We only implemented and evaluated two basic types of active learning strategies in our work, namely the data diversity-based and uncertainty-based strategies, and we are aware there exist other families of AL strategies that could be included in our study, e.g., hybrid approaches.

\bibliography{custom, old_emnlp, label_emnlp}
\clearpage
\appendix

\section{Prompts used for each dataset}
\label{sec:prompt}
Text in [[double brackets]] denotes input data.

\label{sec:appendix}
\subsection{BioMRC}
\begin{lstlisting}[breaklines]
I want you to act as an annotator for a question answering system. You will be given the title and abstract of a biomedical research paper, along with a list of biomedical entities mentioned in the abstract. Your task is to determine which entity should replace the placeholder (XXXX) in the title.

Here's how you should approach this task:

Carefully read the title and abstract of the paper.
Pay close attention to the context in which the placeholder (XXXX) appears in the title.
Review the list of biomedical entities mentioned in the abstract.
Determine which entity from the list best fits the context of the placeholder in the title.
Output only the identifier for the chosen entity (e.g., `@entity1`). Do not output anything else.

<INPUT>:
<title>:
[[TITLE]]
<abstract>:
[[ABSTRACT]]
<entities>:
[[ENTITY]]
<OUTPUT>:
\end{lstlisting}
\subsection{UnfairTOS}
\begin{lstlisting}[breaklines]
I want you to act as an annotator for a Term of Service (ToS) review system. You will be given a piece of a Term of Service. Your job is to determine whether the ToS contains any of the following unfair terms:

Limitation of liability
Unilateral termination
Unilateral change
Content removal
Contract by using
Choice of law
Jurisdiction
Arbitration

If none of the above terms are present, you should output "None".

Here's how you should approach this task:

Carefully read the ToS.
Review the list of unfair terms.
For each unfair term, determine whether it is present in the ToS.
Output only the unfair terms that are present in the ToS. A ToS may have multiple unfair terms. \
You should output all of them, separated by a semicolon (;).
Do not output anything else.

<text>:
[[TEXT]]
<OUTPUT>:
\end{lstlisting}
\subsection{ContractNLI}
\begin{lstlisting}[breaklines]
I want you to act as an annotator for a question answering system. You will be given a contract and a hypothesis. Your task is to determine the hypothesis is contradictory, entailed or neutral to the contract.

Here's how you should approach this task:

Carefully read the contract.
Carefully read the hypothesis.
Determine whether the hypothesis is contradictory, entailed or neutral to the contract.
Output only the label (contradiction, entailment, neutral). Do not output anything else.

<INPUT>:
<premise>:
[[PREMISE]]
<hypothesis>:
[[HYPOTHESIS]]
<OUTPUT>:
\end{lstlisting}
\subsection{FairytaleQA}
\begin{lstlisting}[breaklines]
I want you to act as an annotator for a question generating system. You will be given a story section from a fairytale. Your task is to generate a question that can be answered by the story section. The question should be open (not yes/no) and should be answerable by a child reading the story.


Here's how you should approach this task:

Carefully read the story section.
Determine which question best asks about the story section.
Output only the question. Do not output anything else.

<INPUT>:
<story_section>:
[[STORY_SECTION]]
<OUTPUT>:
\end{lstlisting}

\begin{table*}[t]
    \begin{booktabs}{
        colspec={XXXXXXX},
        width=\linewidth,
        cell{1}{1,2,7}={r=2}{},
        cell{1}{3}={c=4}{},
        cells={m,c}
    }
        \toprule
Dataset     & Metric   & GPT-3.5 &        &        &        & GPT-4  \\
\cmidrule{3-6}
            &          & 0 shot         & 1 shots  & 3 shots  & 10 shots &        \\
\midrule
BioMRC      & Accuracy & 0.4067        & 0.5169 & 0.5040 & 0.4532 & 0.8259 \\
Unfair\_tos & F1       & 0.4201        & 0.3847 & 0.3758 & 0.4206 & 0.4863 \\
ContractNLI & Accuracy & 0.4580        & 0.5990 & 0.5750 & 0.6420 & 0.8240 \\
FairytaleQA & Rouge-L  & 0.3650        & 0.3662 & 0.3834 & 0.3902 & 0.3787 \\
\bottomrule
    \end{booktabs}
    \caption{Large Language Model experiment results.}
    \label{tab:llm}
\end{table*}

\end{document}